\documentclass[letterpaper, 10 pt, conference]{ieeeconf}  

\IEEEoverridecommandlockouts                              

\usepackage{booktabs}
\usepackage{amsmath}
\usepackage{amssymb}
\usepackage{hyperref}
\usepackage{graphicx}
\usepackage{subcaption}
\usepackage{xcolor}
\newcommand{\method}[1]{GCS}

\usepackage{amsmath,amsfonts,bm}

\def\Figref#1{Figure~\ref{#1}}

\def\Secref#1{Section~\ref{#1}}

\def\eqref#1{equation~(\ref{#1})}

\def\Tableref#1{Table~\ref{#1}}

\def\1{\bm{1}}

\DeclareMathAlphabet{\mathsfit}{\encodingdefault}{\sfdefault}{m}{sl}
\SetMathAlphabet{\mathsfit}{bold}{\encodingdefault}{\sfdefault}{bx}{n}

\title{\LARGE \bf
Zero-shot Sim2Real Transfer for Magnet-Based Tactile Sensor on Insertion Tasks
}

\author{
    Beining Han, Abhishek Joshi, Jia Deng \\
    Princeton Unversity
}

\begin{document}

\maketitle
\thispagestyle{empty}
\pagestyle{empty}

\setlength\parindent{10pt}
\begin{abstract}

Tactile sensing is an important sensing modality for robot manipulation. Among different types of tactile sensors, magnet-based sensors, like u-skin, balance well between tactile density, high-durability, and compactness. However, the large sim-to-real gap of tactile sensors prevents robots from acquiring useful tactile-based manipulation skills from simulation data, a recipe that has been successful for achieving complex and sophisticated control policies. Prior work has used binarization techniques to bridge the sim-to-real gap for dexterous in-hand manipulation with magnet-based sensors. However, binarization inherently loses much information that is useful in many other tasks, e.g., insertion. In our work, we propose \method~, a novel sim-to-real technique to learn contact-rich insertion skills with dense, distributed, 3-axes tactile readings from magnet-based tactile sensors. We evaluated our approach on \textit{blind} insertion tasks and show successful zero-shot sim-to-real transfer of RL policies with raw tactile readings as input. Website is: \url{https://princeton-vl.github.io/tactilegcs.github.io/}.

\end{abstract}


\section{Introduction}
\label{sec:intro}

The sense of touch is an important sensing modality for humans. This sense enables humans to perform dexterous tasks such as using tools or detecting subtle changes in contact that can be critical for manipulating complex objects. Several efforts \cite{yuan2017gelsight,  gelslim3, kolamuri2021graspstable, calandra2018graspstable, she2021cable, wilson2023cable, zhao2023fingerslam, suresh2024neuralfeels, qi2023generalinhand, yin2024translationinhand, huang20243dvitac, tomo2016uskin, tomo2018uskin2, bhirangi2024anyskin, lenzexploring, funk2024evetac, akinola2025tacsl} have aimed to replicate such tactile capabilities in robots. In particular, magnet-based tactile sensors such as u-skin \cite{tomo2016uskin, tomo2018uskin2} provide dense tactile information, capturing both normal and shear forces. They are also durable and compact, enabling integration into dexterous hands \cite{paxinigen3hand}.

Furthermore, training reinforcement learning policies in simulation has been an effective technique for learning robot skills \cite{makoviychuk2021isaacgym, kumar2021rma, qi2023generalinhand, yin2023rotatinginhand, he2024agile, chen2023dexori, zhang2024wococo}. The caveat to using simulation is the introduction of the sim-to-real gap. This gap is especially exacerbated for contact-rich tasks involving tactile sensors, making it difficult to transfer policies zero-shot to the real world. Although prior work has explored methods to bridge the sim-to-real gap for magnet-based tactile sensors, such as binarization \cite{qi2023generalinhand, yin2024translationinhand}, these approaches discard dense tactile information and are not suitable for tasks such as blind insertion (\Secref{sec:experiment}). 

\begin{figure*}[ht]
    \centering
    \includegraphics[width=0.95\linewidth]{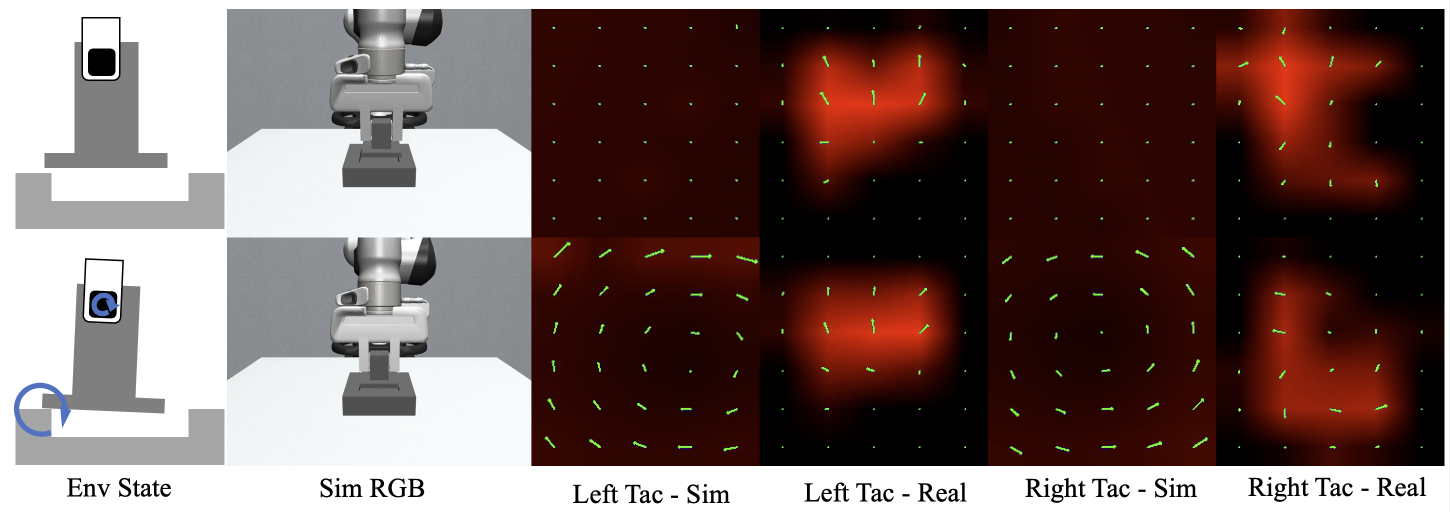}
    \caption{Comparison of simulated $6 \times 5$ tactile readings in MuJoCo and in the real world under the same physical state. The first row is the state where the peg is not in contact with the base. The second row is the state where the peg in contacting the back rim of the base. For the right four figures, the redness represents normal forces and the green arrows represent the shear forces, on each taxel. This visualizes the sim-to-real gap of tactile readings.}
    \label{fig:compare_tactile}
    \vskip -0.25in
\end{figure*}

 In our work, we propose \textbf{\method~}, a novel method to bridge the gap between tactile readings from rigid-body simulation and the real world for magnet-based tactile sensors. \method~ avoids using techniques such as binarization \cite{yin2024translationinhand, qi2023generalinhand} that are prone to information loss. Instead, we opt to bridge the sim-to-real gap directly for \textit{dense, distributed tactile sensor readings} (\Figref{fig:compare_tactile}). Specifically, we show that \method~ enables the zero-shot transfer of RL policies from simulation for blind insertion tasks given raw tactile readings in the form of $6 \times 5 \times 3$ arrays as input.

We identify three key aspects of the sim-to-real gap for magnetic-based tactile sensors: non-uniform contact, contact Poisson Effect, and force scaling difference. For each aspect, we propose a simple yet effective technique to mitigate the gap (\Secref{sec:gcs}). Our techniques involve introducing surfaces with \textbf{G}aussian bumps, applying \textbf{C}onvolutions for the Poisson Effect, and incorporating domain randomization for Force \textbf{S}caling. Moreover, each technique is easy to implement and efficient to compute. \textbf{G} involves a simple randomized initialization of tactile sensor models, and \textbf{S, C} requires randomized data augmentation of the simulated tactile readings. Despite simplicity, we observe that the \method~ readings are qualitatively more similar to their counterparts in the real world (\Figref{fig:gcs_sim}).

In our experiments, we show that \method~ enables \textit{zero-shot} sim-to-real transfer of RL policies on \textit{blind} peg-in-hole insertion tasks, i.e., the robot does not have access to ground truth peg poses. As shown in \Tableref{tab:comparative_eval}, the results suggest that \method~ outperforms all previous sim-to-real techniques and improves the success rate by almost \textbf{ 50\%}. On some tasks, our RL policy is successful in 9 out of 10 real-world trials, while baseline approaches failed completely. Furthermore, we find that each of the techniques that make up \method~ is necessary for the final performance (\Tableref{tab:ablation_study}). To the best of our knowledge, \method~ the first method that bridges the sim-to-real gap on blind insertion tasks of dense, 3-axes, magnet-based tactile sensors.

\section{Related Works}
\label{sec:related_works}

\noindent \textbf{Robot Manipulation with Tactile Sensors.} Tactile sensors are crucial for many manipulation tasks. They provide information of contact between robots and objects, and contact between objects in hand and its surroundings. Moreover, such information cannot be easily revealed from visual information alone. Many prior works used vision-based optical tactile sensors (e.g., Gelsight \cite{yuan2017gelsight,  gelslim3}) in grasping \cite{kolamuri2021graspstable, calandra2018graspstable}, slip detection \cite{li2025modeling}, cable manipulation \cite{she2021cable, wilson2023cable}, cloth manipulation \cite{sunil2023visuotactile}, visual-tactile slam \cite{zhao2023fingerslam, suresh2024neuralfeels}, contact-rich insertion \cite{dong2021tactilerl, kim2022activeinsertion, yu2023mimictouch}, in hand manipulation \cite{qi2023generalinhand} and in exploiting external contact \cite{kim2024texterity, su2024sim2real}. Recently, works have also used piezoelectric and magnet-based sensors \cite{bhirangi2024anyskin, bhirangireskin, pattabiraman2025eflesh} for various manipulation tasks, including visual-tactile in-hand reorientation and translation \cite{yin2023rotatinginhand, yuan2024robot, yin2024translationinhand}. Most importantly, a piezoelectric sensor, i.e., 3D-ViTac \cite{huang20243dvitac, huang2025vtrefine} and a magnet-based sensor, e.g., u-skin \cite{tomo2016uskin, tomo2018uskin2} have been introduced to robot manipulation. These two sensors can provide dense, distributed tactile signals. Works have used these high-dimensional tactile information for complex manipulation tasks via imitation learning on tele-operated demonstrations \cite{huang20243dvitac, wu2024canonical, guzey2023see, guzey2023dexterity, sharma2025selfsupervised}. Unlike these prior work, we focus on bridging the sim-to-real gap of dense, distributed, 3-axes, magnet-based sensors.

\noindent \textbf{Tactile Sensor Sim-to-Real.} There is only limited data involving real-world tactile readings \cite{yang2024bunitouch}, e.s.p., data collected during robot manipulation \cite{kim2024texterity, huang20243dvitac}. In contrast, simulation data are relatively cheap to acquire and work has shown learning dexterous skills with only simulation data \cite{chen2023dexori, qi2023generalinhand, yin2023rotatinginhand}. However, as the mechanisms of the rigid-body physics engines \cite{todorov2012mujoco, makoviychuk2021isaacgym} are different from the real world physics, bridging the large sim-to-real gap for manipulation with tactile sensors remains a challenge. For optical tactile sensors, prior work developed example-based simulators \cite{si2022taxim, wang2022tacto}, differentiable simulators \cite{si2024difftactile}, rigid-body simulators coupled with penalty-based models \cite{akinola2025tacsl, xu2023efficient} and FEM-based simulators \cite{du2024tacipc, chen2024general, li2025taccel}. Notably, \cite{chen2024general} achieved zero-shot sim-to-real transfer on contact-rich insertion tasks \cite{chen2024general} with IPC-based \cite{li2020ipc} simulation. However, these methods are not suitable for magnet-based sensors. In addition, works have also shown success on dexterous in-hand orientation \cite{yin2023rotatinginhand, qi2023generalinhand} and translation \cite{yin2024translationinhand}, with binary tactile signals from magnet-based or piezoelectric sensors. However, binarization loses a lot of information from dense, distributed, 3-axes sensors. For example, in peg-in-hole insertion tasks, binarization does not provide information of contact between the peg and the hole base. In our work, we use the raw tactile reading directly as input to the RL policy and we develop a different technique to bridge the sim-to-real gap of raw tactile reading for dense, distributed, 3-axes, magnet-based sensors.

\section{Method}
\label{sec:method}
\begin{figure*}[t]
    \begin{subfigure}[t]{0.2\textwidth}
        \centering
        \includegraphics[width=\textwidth]{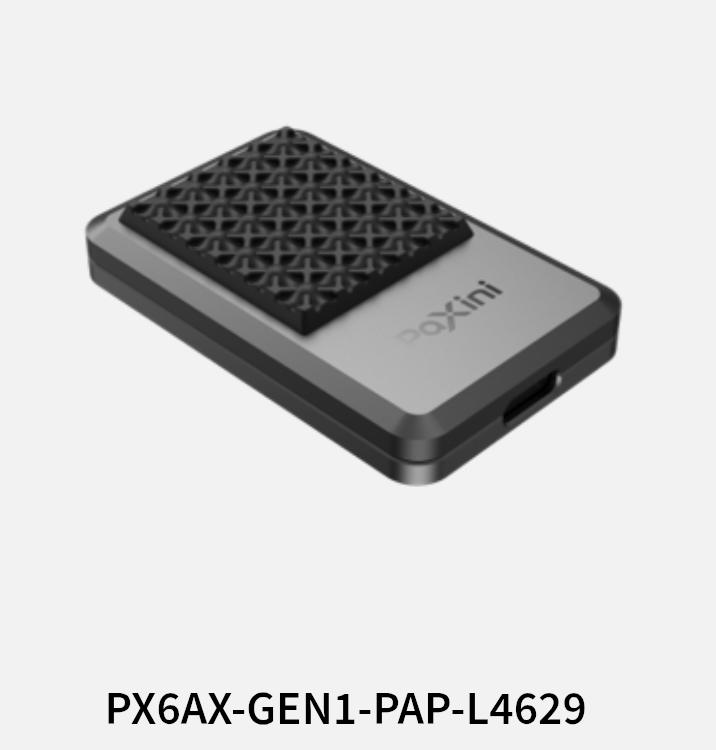}
        \caption{PaXini Sensor}
        \label{fig:paxini_sensor}
    \end{subfigure}
    \begin{subfigure}[t]{0.208\textwidth}
        \centering
        \includegraphics[width=\textwidth]{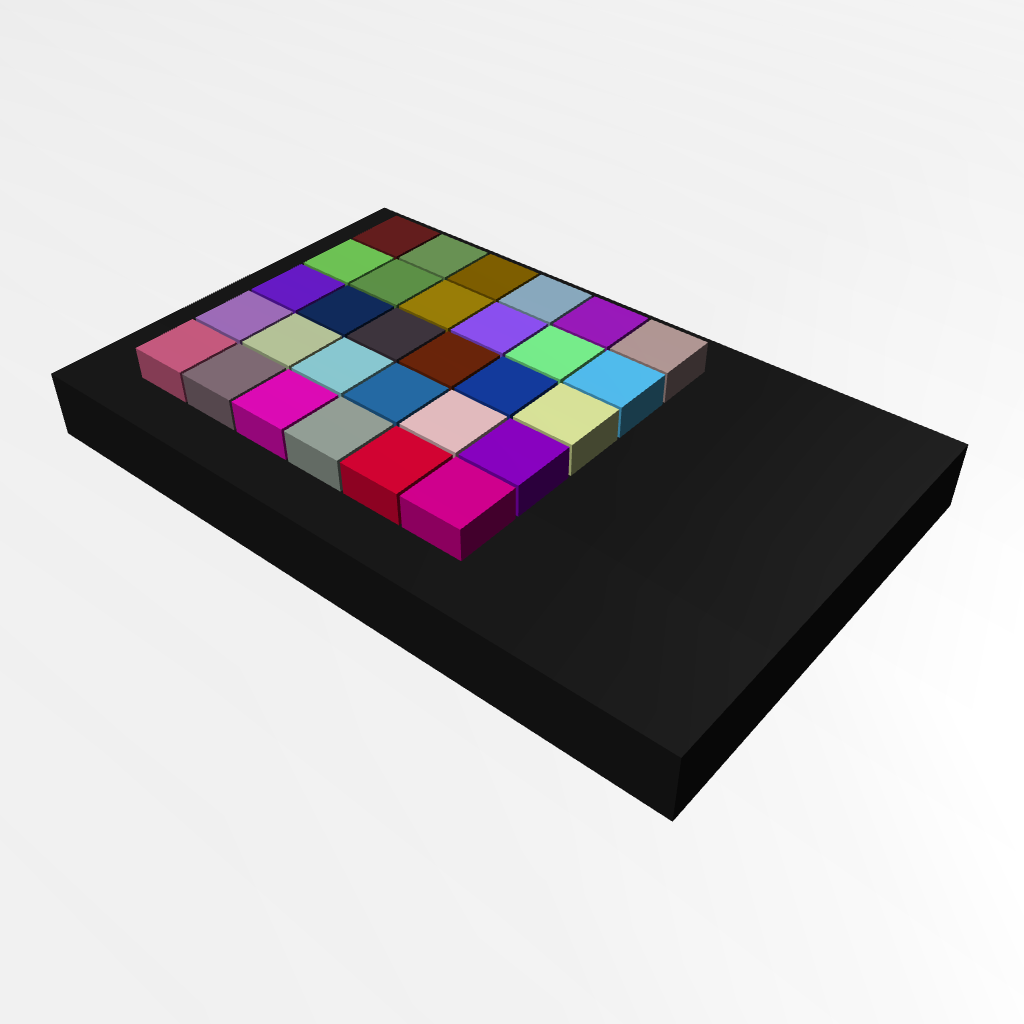}
        \caption{Sensor Model}
        \label{fig:sensor_simulation}
    \end{subfigure}
    \begin{subfigure}[t]{0.2\textwidth}
        \centering
        \includegraphics[width=\textwidth]{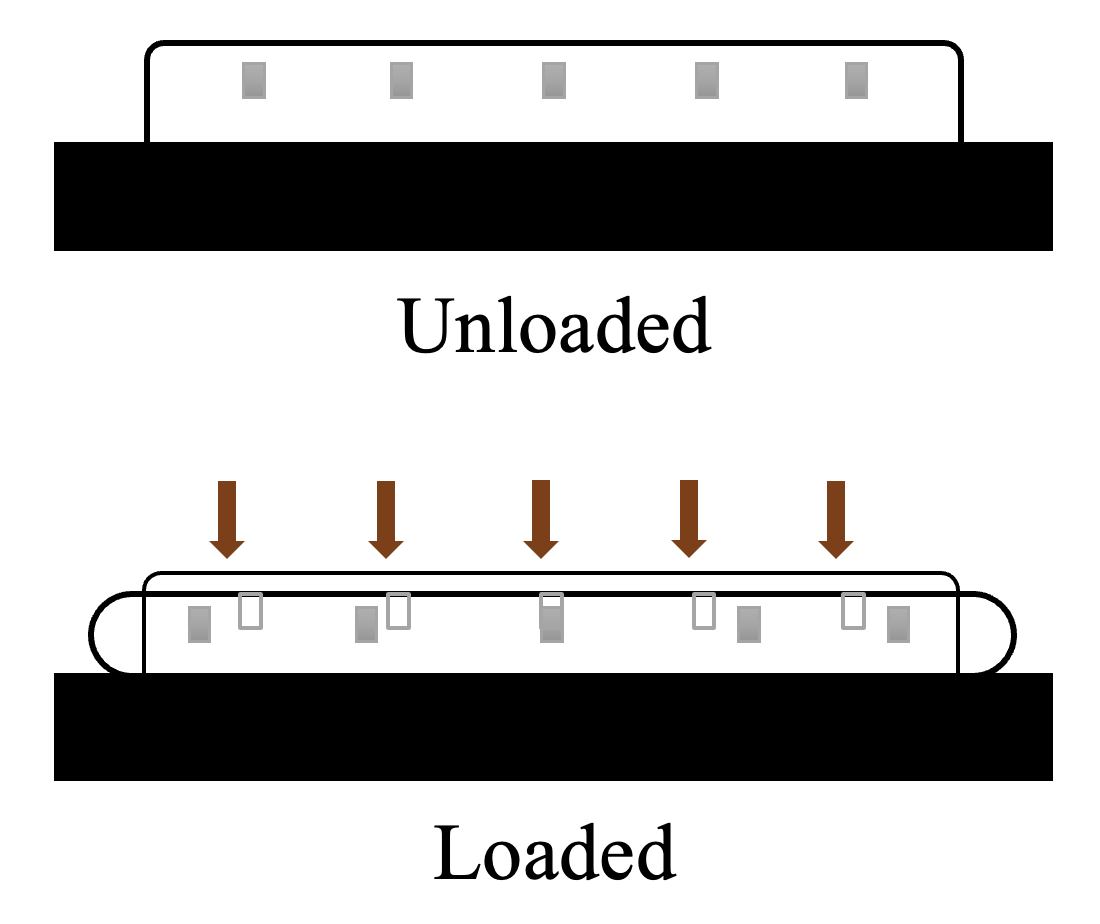}
        \caption{Poisson Effect}
        \label{fig:poisson_effect}
    \end{subfigure}
    \begin{subfigure}[t]{0.35\textwidth}
        \centering
        \includegraphics[width=\textwidth]{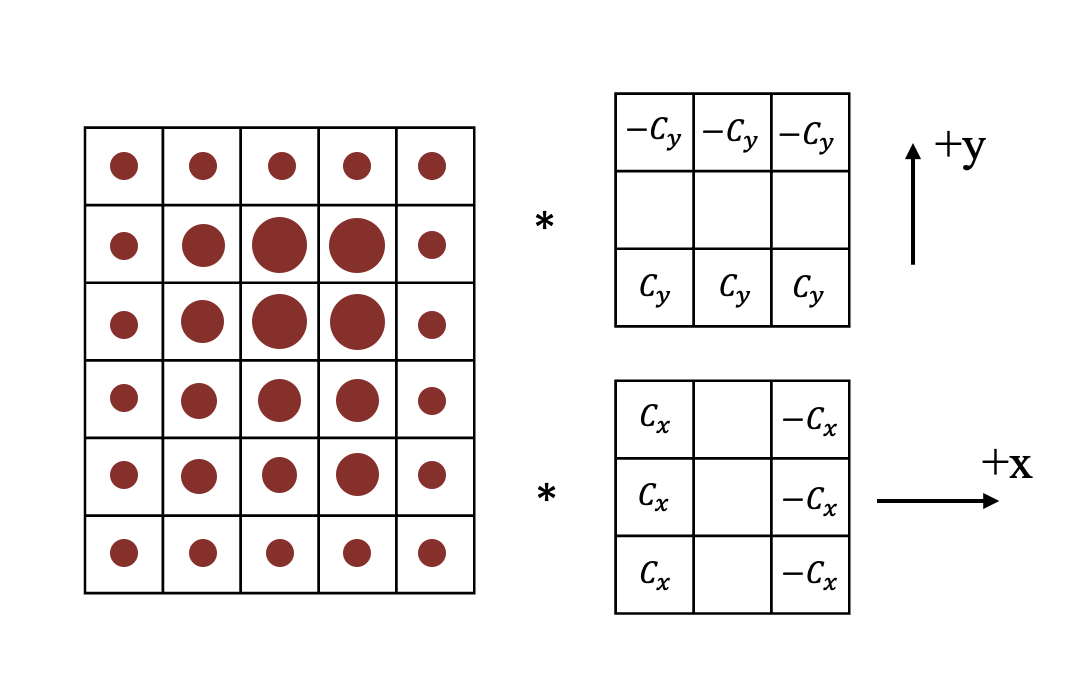}
        \caption{Poisson Convolution}
        \label{fig:poisson_convolution}
    \end{subfigure}
    \caption{(a) Product PX6AX-GEN1-PAP-L4629 from PaXini Tech, figure from \cite{paxinigen1}. (b) MJCF visualization of the tactile sensor model in simulation. We use $6\times5$ small cubes instead of a single cube. (c) Illustration of Poisson Effect on the tactile sensor in real world. Gray cubes represent the small magnets injected in the gel \cite{tomo2016uskin}. (d) Illustration of convolution technique to approximate Poisson Effect noise in real sensor readings. Red dots represent the normal force on each taxel, which is convolved with poisson-effect kernels in both directions. $c_{x, y}$ are hyper-parameters that control the noise scale.}
    \vskip -0.1in
\end{figure*}

In our work, we use the dense, distributed, 3-axes, magnet-based tactile sensor: PX6AX-GEN1-PAP-L4629 (\Figref{fig:paxini_sensor}) from PaXini Tech \cite{paxinigen1}. It is a commercial version of u-skin sensor \cite{tomo2016uskin}. This sensor is more cost-effective compared to products from other manufacturers at roughly $\$200$ per pad. Moreover, these sensors are densely distributed, providing a 3-axes force reading of a $12\times10$ taxel grid on the $26.1mm \times 22.6mm $ surface area. This is significantly denser than most counterparts. Here, we down-sample the reading to $6\times5$ by $2\times2$ sum pooling, in order to improve the simulation speed in RL training.

To simulate tactile sensors, we extend Robosuite \cite{zhu2020robosuite} which extends the MuJoCo \cite{todorov2012mujoco} physics engine. MuJoCo features soft-contact and explicit dynamics integration. Compared to other common simulators, e.g., Isaac-Gym \cite{makoviychuk2021isaacgym} or Bullet \cite{coumans2015bullet}, this soft-contact modeling produces more consistent and stable force readings. Similar to \cite{sferrazza2024power}, we use the Touch Grid plugin provided by MuJoCo to get the aggregate contact forces in each taxel area. In addition, as shown in \Figref{fig:sensor_simulation}, instead of using a single cube as the contact pad of the sensor, we discretize the contact pad with small cubes of $4.6mm \times 4.6mm \times 1mm$ (W$\times$H$\times$D). This produces dense contact force readings, while a single contact cube will result in a sparse contact force due to the underlying collision detection.

\begin{table*}[htb]
  \centering
  \caption{Domain Randomization (DR) distribution of key parameters in our GCS method.}
  \begin{tabular}{ccccc}
    \toprule
    Parameters & Axes Force $\alpha_k$ & Taxel Force $\beta_{ij}$ & Convolution $c_{x,y}$ & Deviation $s_{x, y}$ \\
    \midrule
    DR Range & $\mathcal{U}(0.5, 1.5)$ & $\mathcal{N}(1.0, 0.25),  20\%$ Dropout & $\mathcal{U}(0.1, 0.3)$ & $\mathcal{U}(1.0, 3.0)$ \\
    \bottomrule
 \end{tabular}
 \label{tab:domain_randomization}
 \vskip -0.15in
\end{table*}

\Figref{fig:compare_tactile} compares tactile readings in MuJoCo with readings from real world sensors under the same physical state. In the first row, the robot holds the cubic peg handle with no contact to its external environment. In the second row, the robot holds the same handle while the peg is in contact with the back rim of the square hole. The degree of redness reflects the amount of normal force on each taxel, while the green arrows represent the shear force direction and its scale. It is evident that there exists a significant sim-to-real gap in tactile sensor readings. We identify that the gap comes from three main aspects:
\begin{enumerate}[leftmargin=0.2in, itemsep=0pt, topsep=0pt, partopsep=0pt]
    \item \textbf{Non-uniform Contact}. Due to a manufacturing defect, the actual surface gel is uneven and the injected small magnets \cite{tomo2016uskin} are not completely symmetric in terms of magnetic field. Therefore, we observe non-uniform activation of the dense, distributed Hall Effect sensors. That is despite flat contact surfaces, we still observe non-uniform normal force readings. Moreover, such non-uniformity is inconsistent between different products. We note that left and right tactile readings may be different while their contact surfaces and gripper forces are symmetric.
    \item \textbf{Contact Poisson Effect}. There exists a significant non-zero shear force even when the peg is not in contact with the environment. We find that this results from the Poisson Effect of the contact gel, which is illustrated in \Figref{fig:poisson_effect}. This effect also leads to crosstalk among adjacent taxels \cite{tomo2016uskin, tomo2018uskin2}. We find that the shear magnet displacement is in proportion with the scale of the loaded force in proximity.
    \item \textbf{Different Force Scales}. The absolute scale of real-world sensor force differs from the scale in simulation. This comes from the difference between the physics engine and real-world physics. In MuJoCo, the contact force is determined by minimizing the next-step kinetic energy \cite{todorov2012mujoco}, which is subject to hyperparameters, like impulse regularization scaling and error reduction time constant. Different physical parameters will result in a different contact force scale in simulation. In addition, we also find that the real sensor reading is quite inaccurate.
\end{enumerate}

\subsection{GCS: Bridge Sim-to-Real Gap}
\label{sec:gcs}

Prior works bridge the sim-to-real gap via binarization techniques \cite{qi2023generalinhand, yin2023rotatinginhand, yin2024translationinhand}. However, this technique loses information that is crucial for many tasks. For example, for insertion tasks (\Secref{sec:experiment}), binarization of taxel forces does not provide any information on the contact force between the peg and the environment. In our work, we aim to bridge the sim-to-real gap of \textit{raw tactile readings} of dense, distributed, 3-axes tactile forces, as it is the most general form of tactile input. To this end, we propose three simple yet effective techniques to address the above gaps.

\noindent \textbf{1. Surface with Gaussian Bumps.} To simulate the effect of non-uniform contact of the taxels, we randomize the depth of the small cubes on the surface pad in simulation. Specifically speaking, we randomly select a 2-dimensional Gaussian distribution, i.e., mean $(g_x, g_y) \in \mathcal{G} = [0, 5] \times [0, 6]$ and standard deviation $(s_x, s_y)$. Then, given the coordinate of each taxel $(i, j) \in \mathcal{G}$, we denote $D((i,j), (g_x, g_y)) = \sqrt{(g_x - i)^2 / s_x^2 + (g_y - j)^2 / s_y^2}$. The depth $h(i,j)$ of taxel cube $(i,j)$ is
\begin{align*}
h(i,j) = h_{\text{min}} + \frac{D((i,j), (g_x, g_y))}{\max_{(i,j) \in \mathcal{G}}{D((i,j), (g_x, g_y))}}(h_{\text{max}} - h_{\text{min}}).
\end{align*}
Here, $h_{\text{max}}$ and $h_{\text{min}}$ denote the maximum and minimum cube depths, which are chosen as hyperparameters. Additionally, based on observations from real-world tactile readings, we reduce the depth at the boundary of the surface taxel grid. (\Figref{fig:compare_tactile}).

\noindent \textbf{2. Convolution for Poisson Effect.} High fidelity simulation of deformable objects require FEM-based approaches \cite{li2020ipc}, which are computationally expensive. Here, we approximate the Poisson Effect noise via a simple approach. Observing that the scale of the shear force is proportional to the normal force in its surroundings, we approximate the Poisson Effect shear displacement by convolving the normal forces tensor with poisson-effect kernels, illustrated in \Figref{fig:poisson_convolution}. The poisson-effect scale is determined by parameters $c_x, c_y$ respectively for each direction.

\noindent \textbf{3. Domain Randomization of Force Scaling.} To address the sim-to-real gap from inappropriate physical parameters and inaccurate sensor readings, we randomize the force scaling factors in each episode during training. Here, we randomize the scaling for each axis of all taxels, i.e.,  $\alpha_k, k \in [x, y, z]$, and independently randomize the scale of each taxel $\beta_{ij}, (i,j) \in \mathcal{G}$. Namely, each taxel reading $f_{k}(i,j) = \hat{f}_k(i,j) * \alpha_k * \beta_{ij}$, where $\hat{f}_k(i,j)$ is the raw force reading from the physics engine for taxel $(i,j)$ in axis $k$. 

\begin{figure*}[ht]
    \centering
    \includegraphics[width=0.95\linewidth]{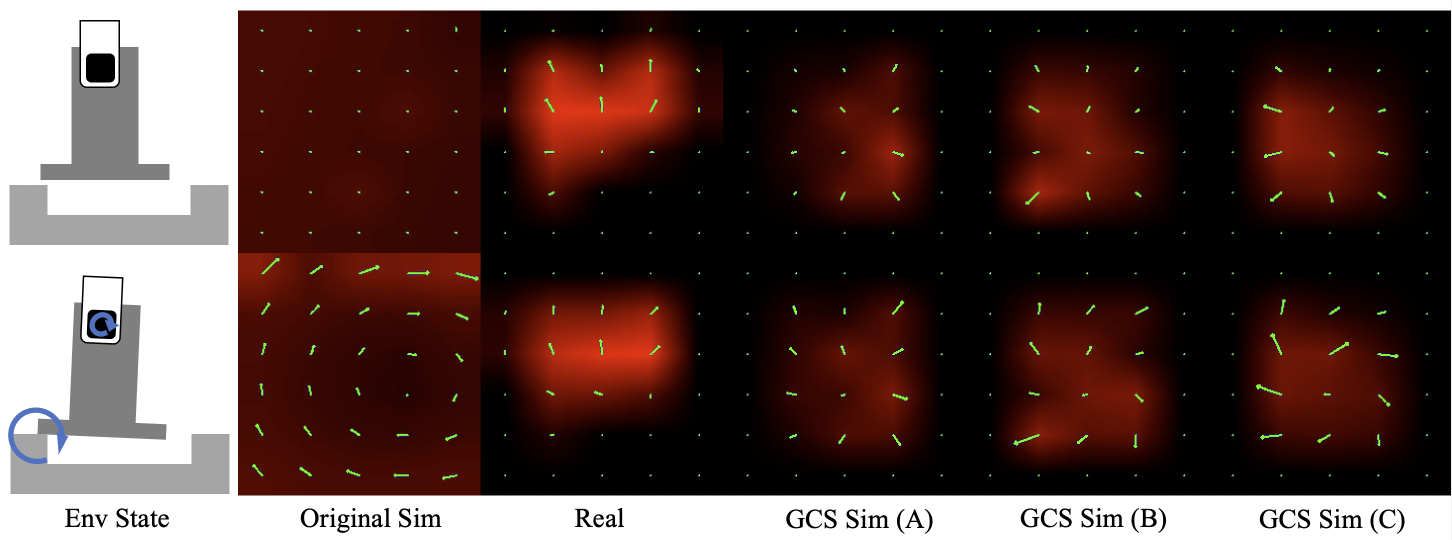}
    \caption{The 3 rightmost columns (A-C) show GCS tactile readings from the left finger under 3 random parameters sampled from \Tableref{tab:domain_randomization}. We compare them with the real-world tactile readings and the original raw simulation tactile readings under the same physical states as in \Figref{fig:compare_tactile}.}
    \label{fig:gcs_sim}
    \vskip -0.2in
\end{figure*}

\Tableref{tab:domain_randomization}
summarizes the randomization range of all factors in our pipeline. \Figref{fig:gcs_sim} illustrates the revised tactile readings with \method~, in the same physical states as \Figref{fig:compare_tactile}. In \Secref{sec:experiment}, we show that the proposed method can bridge the sim-to-real gap for \textit{blind} insertion tasks, allowing zero-shot transfer of an RL policy with raw tactile sensor readings as input.

\section{Experiment}
\label{sec:experiment}

In this section, we test \method~ on transferring simulated tactile information to the real world for insertion tasks. Here, we aim to answer the following questions.
\begin{enumerate}[leftmargin=0.2in, itemsep=-0.5pt, topsep=-0.5pt, partopsep=-1pt]
    \item How does \method~ perform compared to other sim-to-real techniques in previous work?
    \item How does each technique in \method~ contribute to real-world performance?
\end{enumerate}

\subsection{Experiment Setup}
\label{sec:exp_setup}

\begin{figure*}[ht]
    \centering
    \vskip -1.95in
    \includegraphics[width=\linewidth]{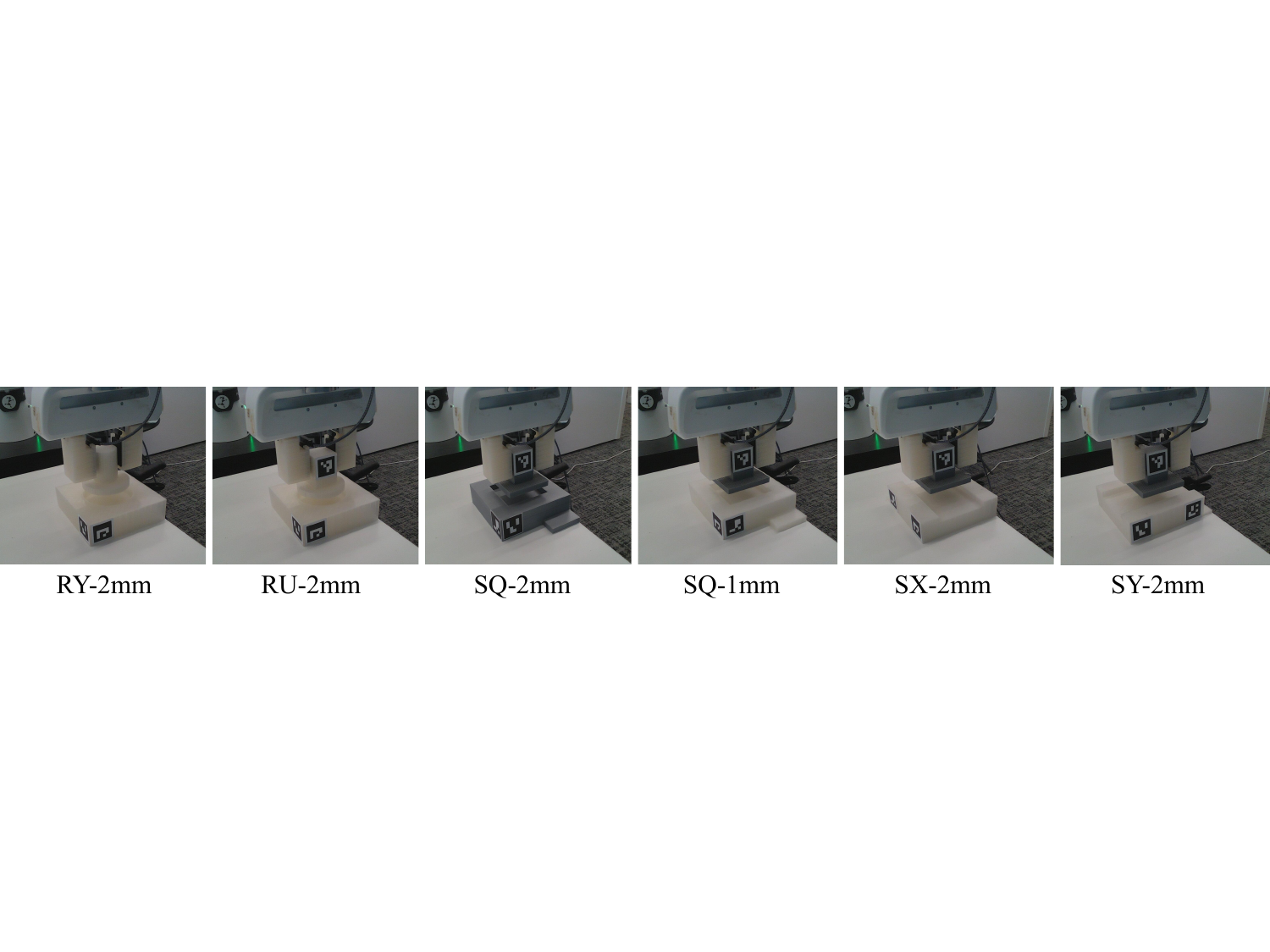}
    \vskip -1.95in
    \caption{Illustration of all 6 \textit{blind} insertion tasks in our experiment.}
    \label{fig:task_illustration}
    \vskip -0.15in
\end{figure*}

\noindent \textbf{Task.} We use the following 6 \textit{blind} insertion tasks in our experiment (\Figref{fig:task_illustration}). 

\begin{enumerate}[leftmargin=0.2in, itemsep=-1pt, topsep=-1pt, partopsep=-1pt]
    \item[-] \textbf{RY-2mm}: Round peg with cylinder handle, and 2mm clearance round hole.
    \item[-] \textbf{RU-2mm}: Round peg with cubic handle, and 2mm clearance round hole.
    \item[-] \textbf{SQ-2mm}: Square peg with cubic handle, and 2mm clearance square hole.
    \item[-] \textbf{SQ-1mm}: Square peg with cubic handle, and 1mm clearance square hole.
    \item[-] \textbf{SX-2mm}: Square peg with cubic handle, and 2mm clearance groove along the x-axis. In addition, the hole pose is also blind to the robot.
    \item[-] \textbf{SY-2mm}: Square peg with cubic handle, and 2mm clearance groove along the y-axis. In addition, the hole pose is also blind to the robot.
\end{enumerate}

For each task, the robot does not know the pose of the peg, and the peg is randomly placed between its fingers during initialization. Consequently, the robot is required to infer contact between the peg and the hole base via tactile inputs. In our experiment, the robot hand is initialized such that the bottom of the peg is located within $[-1.5cm, 1.5cm] \times [1.5cm, 1.5cm] \times [0.5cm, 2cm]$ (XYZ) of the center of the hole. Please refer to the appendix for task details.

\noindent \textbf{Hardware.} We 3d-printed custom fingers to attach sensors to the Franka Research 3 Gripper (\Figref{fig:task_illustration}). We place Realsense L515 in front of the scene to capture object poses, which are labeled with Aruco markers. All poses are computed relative to the robot base frame with camera calibration. Notice that in addition to peg poses, base poses in SX-2mm and SY-2mm are also not available to the robot. Here, we put markers for tracking and task initialization. We use Deoxys \cite{zhu2022viola} to run the RL policy with an Operation Space Controller (OSC) at 20Hz on a real robot.

\noindent \textbf{RL Policy.} For all policies, the input includes the relative pose from the robot fingertip to the hole base $p_t$ (zeros for SX-2mm, SY-2mm tasks). For \method~ and baselines (\Secref{sec:comparative_eval}) with tactile inputs, we use a history of $k$ tactile readings, where $k$ is a hyperparameter for each method. In \method~, the tactile input $o_t$ is a $6k \times 6 \times 5$ ($C \times H \times W$) tensor. The policy network first encodes the tactile reading with a CNN to a 128-dim latent vector $t_t = T(o_t)$ and encodes $h_t = E(p_t)$ to a 128-dim vector. Then, we feed them into a 2-layer MLP action head $a_t = \pi(E(p_t), T(o_t))$. Here, we use asymmetric SAC \cite{haarnoja2018sac} to train the RL policy in Robosuite \cite{zhu2020robosuite}, implemented based on StableBaseline3 \cite{stablebaseline3}. The critic network takes ground truth pose of objects and fingertip poses as input without tactile readings. For all tasks, the reward is the sum of the reaching reward, engagement reward, and success bonus,  which is similar to \cite{tang2023industreal}. Please refer to the appendix for RL training details.

\begin{table*}[h]
  \centering
  \caption{Comparative evaluation on success rate of baselines and ours \method~ in simulation and real world. NT: No Tactile; TF: Total Force; TB: Taxel Binarization; TD: Taxel Direction; DT: Direct Transfer.}
  \resizebox{0.85\linewidth}{!}{\begin{tabular}{cccccccc}
    \toprule
    Method & RY-2mm & RU-2mm & SQ-2mm & SQ-1mm & SX-2mm & SY-2mm & Avg \\
    \midrule
    \color{gray}{NT (Sim)} & \color{gray}{0.94} & \color{gray}{0.82} & \color{gray}{0.6} & \color{gray}{0.56} & \color{gray}{0.42} & \color{gray}{0.2} & \color{gray}{0.59} \\
    NT (Real) & 0.4 & 0.4 & 0.0 & 0.1 & 0.0 & 0.0 & 0.15 \\
    \midrule
    \color{gray}{TF (Sim)} & \color{gray}{1.0} & \color{gray}{0.98} & \color{gray}{0.8} & \color{gray}{0.0} & \color{gray}{1.0} & \color{gray}{0.74} & \color{gray}{0.75} \\
    TF (Real) & 0.4 & 0.3 & 0.3 & 0.0 & 0.4 & 0.1 & 0.25 \\
    \midrule
    \color{gray}{TB (Sim)} & \color{gray}{1.0} & \color{gray}{0.82} & \color{gray}{0.62} & \color{gray}{0.36} & \color{gray}{0.84} & \color{gray}{0.1} & \color{gray}{0.62} \\
    TB (Real) & 0.5 & 0.3 & 0.4 & 0.1 & 0.5 & 0.2 & 0.33 \\
    \midrule
        \color{gray}{TD (Sim)} & \color{gray}{1.0} & \color{gray}{0.86} & \color{gray}{0.9} & \color{gray}{0.88} & \color{gray}{0.88} & \color{gray}{0.80} & \color{gray}{0.88} \\
    TD (Real) & 0.1 & 0.4 & 0.0 & 0.2 & 0.2 & 0.2 & 0.18 \\
    \midrule
    \color{gray}{DT (Sim)} & \color{gray}{1.0} & \color{gray}{0.96} & \color{gray}{0.84} & \color{gray}{0.94} & \color{gray}{0.96} & \color{gray}{0.84} & \color{gray}{0.92} \\
    DT (Real) & 0.6 & 0.2 & 0.1 & 0.3 & 0.4 & 0.1 & 0.28 \\
    \midrule
    \color{gray}{GCS (Sim)} & \color{gray}{1.0} & \color{gray}{0.9} & \color{gray}{0.84} & \color{gray}{0.98} & \color{gray}{0.98} & \color{gray}{0.76} & \color{gray}{0.91} \\
    GCS (Real) & \textbf{0.9} & \textbf{0.8} & \textbf{0.9} & \textbf{0.6} & \textbf{0.9} & \textbf{0.7} & \textbf{0.8} \\
    \bottomrule
 \end{tabular}}
 \label{tab:comparative_eval}
 \vskip -0.2in
\end{table*}

\subsection{Comparative Evaluation}
\label{sec:comparative_eval}
We compare \method~ with the following baselines, Similar to \method~, each baseline is trained with asymmetric SAC (\Secref{sec:exp_setup}). Implementation details are listed in the appendix.

\begin{enumerate}[leftmargin=0.2in, itemsep=0pt, topsep=0pt, partopsep=-1pt]
    \item \textbf{No Tactile (NT).} The policy $\pi(p_t)$ is a 2-layer MLP that does not use tactile information.
    \item \textbf{Total Force (TF).} Total contact force $f_t$ on each surface pad is computed by summing all taxel readings. The force input $f_t$ is a $6k$-dim vector, which is concatenated with $p_t$, i.e., $\pi(p_t, f_t)$.
    \item \textbf{Taxel Binarization (TB).} Similar as \cite{yin2023rotatinginhand, qi2023generalinhand}, we binarize the taxel reading with a 0.1N threshold. The tactile input is a $2k \times 6 \times 5$ tensor of 0 and 1. We encode with a similar CNN.
    \item \textbf{Taxel Direction (TD).} Inspired by \cite{yin2024translationinhand}, we binarize each taxel reading in each axis with 0.1N threshold, together with the sign to represent the direction. Thus, the tactile input is a $6k \times 6 \times 5$ tensor of value $[-1, 0, 1]$, encoded with a similar CNN.
    \item \textbf{Direct Transfer (DT).} We use the tactile reading from the simulator (\Figref{fig:compare_tactile}) directly, which is a $6k \times 6 \times 5$ tensor, and is encoded with a similar CNN encoder.
\end{enumerate}

For real-world evaluation, we test each method with 10 randomly initialized trials. In simulation, we test with 50 trials. The result is shown in \Tableref{tab:comparative_eval}. 

We find that baselines cannot produce good policies that transfer well to the real world. For NT, it has a $<$60\% success rate on square peg tasks even in simulation. When transferred to the real world, it has a $<$10\% success rate. We find that this comes from both pose tracking noise and OSC dynamics gap due to friction in physical motors. NT performs better on round peg tasks, as the round peg is rotation symmetric and thus is more compliant and tolerant to collision. In all, the low performance of NT suggests that \textit{blind} insertion tasks require tactile information.

TF has a higher success rate in simulation than NT. However, as discussed in \Secref{sec:method}, the real-world reading is quite noisy. We observe that the total force on the pad can even point upward when there is no external contact. Consequently, the policy cannot perform well in real world.

Though binarization techniques like TB are useful for in-hand orientation tasks \cite{qi2023generalinhand, yin2023rotatinginhand} and TD is effective for in-hand translation \cite{yin2024translationinhand}, they are not useful for insertion tasks. In fact, the policy in TB learns to change the peg pose via extrinsic contact, in order to infer the contact with the hole base. However, such policy is very hard to transfer to real world. Additionally, in TD, the real-world shear force noise makes the feature in TD non-transferrable to the real-world rollouts. Thus, TD has a quite large sim-to-real performance gap.

With dense, distributed, 3-axes force, the policy can get $>90\%$ success rate in simulation (DT). However, owing to large sim-to-real gap (\Figref{fig:compare_tactile}), the real-world success rate is low due to distribution shift.

We observe that \method~ outperforms all baselines by a 50\% margin on average. In many tasks such as RY-2mm, SQ-2mm, and SX-2mm, it has a real-world success rate of 90\% with barely any performance drop than in the simulation. The largest sim-to-real gap occurs in the SQ-1mm case, which is the hardest task owing to the small clearance. On SQ-1mm, one needs fine-grained and compliant control. Small action error, potentially from sim-to-real tactile gap, will cause the peg to miss the hole in the next step.

\subsection{Ablation Study}
\label{sec:ablation}

\begin{figure*}[ht]
    \centering
    \includegraphics[width=0.85\linewidth]{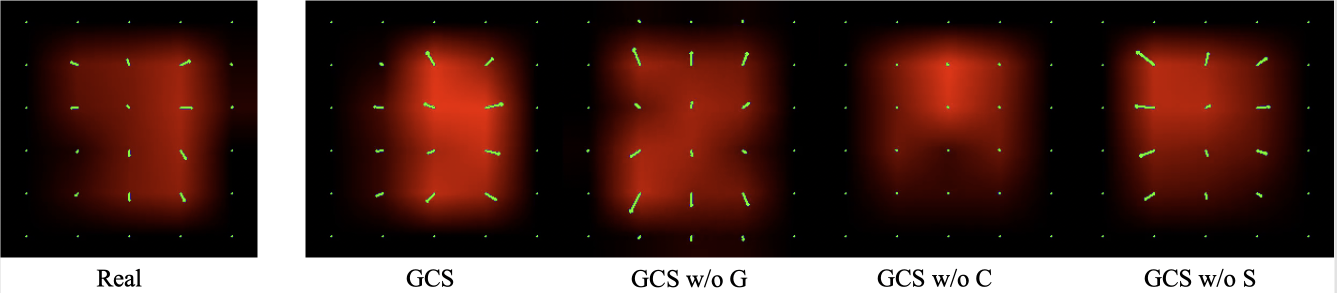}
    \vskip -0.1in
    \caption{Visualization of ablation's left tactile readings in the no contact state. Noticing that readings in \method~ and ablations are randomized in training.}
    \label{fig:ablation_visualization}
    \vskip -0.1in
\end{figure*}

\begin{figure*}[ht]
    \centering \includegraphics[width=0.85\linewidth]{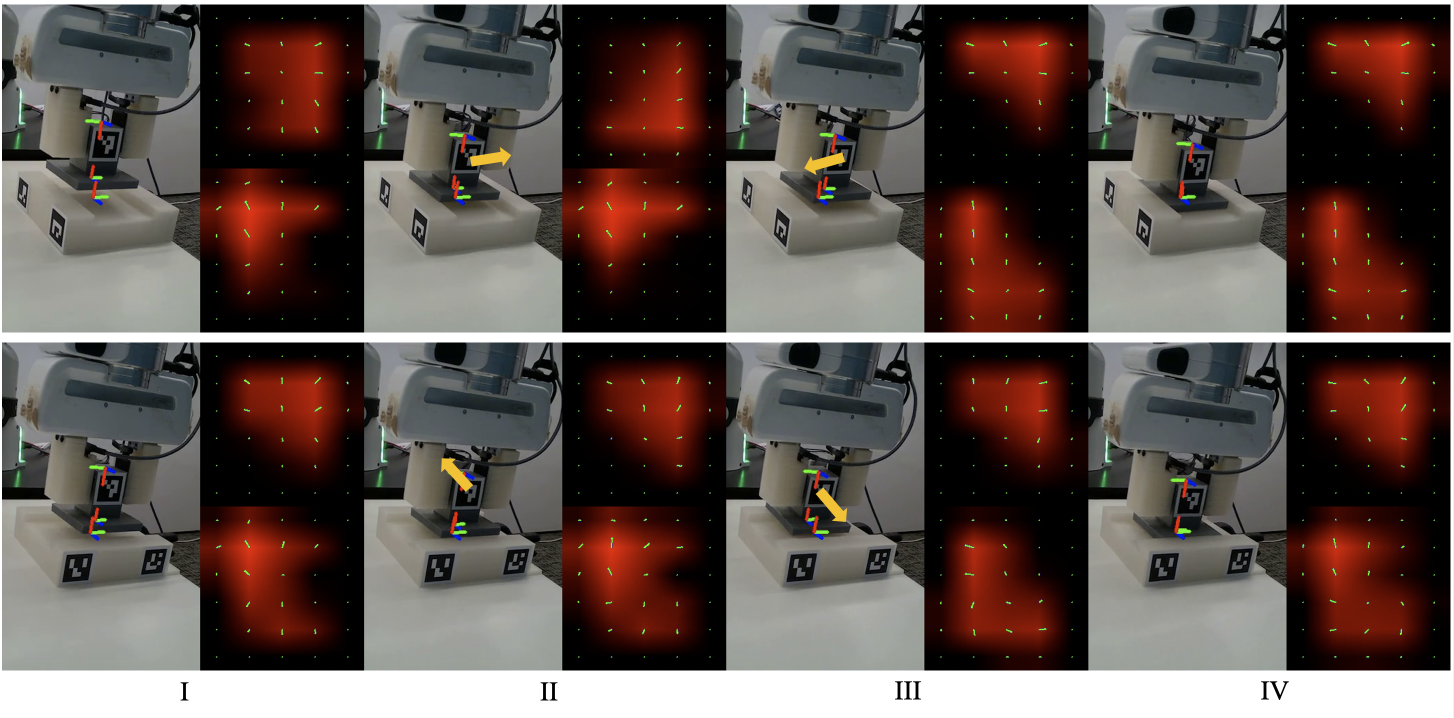}
    \vskip -0.1in
    \caption{Examples of policy rollout in SX-2mm (first row) and SY-2mm (second row). Stage I corresponds to the initial state, and Stage IV corresponds to the success state. Stage II and Stage III corresponds to the critical contact states that the policy infers the relative position from (\Secref{sec:policy_interpretation}). The yellow arrow represents the action of the current step, which can be interpreted with the tactile readings.}
    \label{fig:task_rollout}
    \vskip -0.3in
\end{figure*}

Here, we demonstrate the necessity of all \textit{G, C, S} technique to bridge the sim-to-real gap.
\Tableref{tab:ablation_study} shows the real-world success rate of the ablation. In simulation, the success rate is $>$80\% for all ablations. Clearly, results show that all three \textit{G,C,S} techniques are necessary. This is especially the case in SX-2mm and SY-2mm tasks, where it only uses tactile reading as input. Consequently, distribution mismatch between simulated and real-world tactile readings will result in the failure of the contact-rich insertion task. \Figref{fig:ablation_visualization} visualizes the tactile reading in simulation of different ablations. With all \textit{G,C,S} techniques, the reading looks more similar to real world. 

\begin{table}[h]
  \centering
  \vskip -0.1in
  \caption{Ablation study of \method~ on SX-2mm, SY-2mm, and SQ-2mm tasks. In each ablation, we remove one technique in \Secref{sec:gcs}.}
  \begin{tabular}{cccccccc}
    \toprule
     & GCS & GCS w/o G & GCS w/o S & GCS w/o C \\
    \midrule
    SX-2mm & 0.9 & 0.4 & 0.3 & 0.3 \\
    SY-2mm & 0.7 & 0.0 & 0.6 & 0.2 \\
    SQ-2mm & 0.9 & 0.9 & 0.8 & 0.6 \\
    \bottomrule
 \end{tabular}
 \vskip -0.2in
 \label{tab:ablation_study}
\end{table}

\subsection{Policy Interpretation}
\label{sec:policy_interpretation}

To better understand how the policy trained in simulation transfers to real world, we visualize trajectories in SX-2mm and SY-2mm tasks, in \Figref{fig:task_rollout}. In SY-2mm, it detects the relative position between the peg and the hole by the change of the shear forces. In stage II, the peg contacted with the front rim and thus shear force markers on the left finger changed counterclockwise and clockwise on the right finger sensor. Consequently, the policy commanded the hand to move backward. However, it overshoot and contacted the back rim (stage III). Again, it detected the opposite change on both sensors and thus moved forward. This time, it managed to insert the peg in the groove.

For SX-2mm, it detects the relative position by the change of distribution of normal force on the sensor pads. In stage II, it contacted the left rim and the normal force shifted downward on the left sensor, owing to the external torque. The policy then commanded it to move right. In stage III, it contacted with the right rim and the normal force on the right pad shifted downward this time. Then it moved left in fine-grained steps until the peg was inserted into the groove.

\section{Conclusion}
\label{sec:conclusion}

Dense, distributed, 3-axes, magnet-based tactile sensor (i.e., u-skin) balances well between durability, tactile density, and compactness. In our work, we try to bridge the sim-to-real gap of these sensors. Here, we identify three aspects of the sim-to-real gap: non-uniform contact, poisson effect, and scale difference, and propose a novel \method~ method to mitigate these gaps. We evaluate our approach on \textit{blind} insertion tasks and show that \method~ improves the real-world success rate by 50\% over baselines. Our work creates new possibilities to future sim-to-real learning of visual-tactile manipulation skills with magnet-based sensors.

\section*{ACKNOWLEDGMENT}

This work was partially supported by the National Science Foundation. The authors also thank Haozhi Qi for discussions on the project.

\bibliographystyle{bibtex/IEEEtran}
\bibliography{bibtex/IEEEabrv,bibtex/IEEEexample}

\vskip -0.35in
\appendix

\noindent \textbf{Task Setup.} We use OSC in all tasks. For RY-2mm, RU-2mm, SQ-2mm, SQ-1mm, the policy network outputs 3-dim position delta (XYZ) in $[-1, 1]$, scaled with 1cm, except for SQ-1mm, which is scaled with 5mm. Noticing that we freeze the rotation in OSC to simplify the tasks. For SX-2mm, we further freeze X-axis and for SY-2mm, we freeze Y-axis.

At initialization, the robot grasps the peg and moves the peg to the top of the hole. We put the bottom of the peg randomly within $[-1.5cm, 1.5cm] \times [1.5cm, 1.5cm] \times [0.5cm, 2cm]$ (XYZ) from the center of the hole. 

We set a maximal 120 execution steps for SX-2mm and SY-2mm, and 200 steps for the rest of the tasks. The rollout is considered successful if the peg is successfully inserted into the hole within the trajectory length limit.

\noindent \textbf{Simulation Reward.} The RL reward for all tasks is defined as the weighted sum of the following 4 terms:
\begin{enumerate}[leftmargin=0.2in, itemsep=-0.5pt, topsep=-1pt, partopsep=-1pt]
    \item[-] \textit{Reaching Reward.} We compute the average distance $d$ of keypoints of the peg to correspondent points on base. The reaching reward is $r_{\text{reach}} = \frac{1}{10 * d + \epsilon}$. We set weight $w_{\text{reach}}$ to be 1.
    \item[-] \textit{Engagement Reward.} Given the engagement distance $h$ (the distance between the bottom of peg to the base bottom), if the peg is engaged with the hole, then we have $r_{\text{engage}} = \frac{1}{10 * h + \epsilon}$, otherwise $r_{\text{engage}} = 0$. We set weight $w_{\text{weight}}$ to be 50.
    \item[-] \textit{Success Bonus.} The success bonus $r_{\text{succ}}$ if $h < h_{\text{th}}$, where $h_{\text{th}}$ is a threshold value. We set weight $w_{\text{succ}}$ to be 200.
    \item[-] \textit{Action Penalty.} We penalize the action scale with $r_{\text{action}} = - \parallel a \parallel^2$. We set weight to be either $w_{\text{action}}$ to be 0.1.
\end{enumerate}

\noindent \textbf{RL Policy.} We select the optimal $k$ in $\{1, 5, 10, 20\}$ for all baselines. We choose k$=$1 for TF, k$=$10 for TB, TD and DT. For \method~, we choose k$=$20 for all tasks except SY-2mm, which we find k$=$10 works better. In \Secref{sec:ablation}, \method~ ablations use the same $k$ as \method~. We also find that a short history has lower performance. On SQ-1mm,\method~ using k$=$5 has only 30\% success rate.

The tactile encoder $T$ in TB, TD, DT, and \method~ is a 4-layer CNN with BatchNorm and ReLU activation and the encoder for $p_t$ is a single-layer MLP. All policy heads are 256-dim, 2-layer MLPs. For the critic in SAC, it is a 2-layer MLP that takes ground truth relative pose from peg to base and end-effector to base as inputs. Here, we use SAC. the off-policy algorithm, implemented in StableBaseline3 \cite{stablebaseline3}, for sample efficiency. Robosuite \cite{zhu2020robosuite} is cpu-based and is relatively slow in simulation speed. Here, we use the default hyper-parameters in StableBaseline3 of SAC on all tasks.

\noindent \textbf{Randomization Range.} We set the randomization range of \method~ (\Tableref{tab:domain_randomization}) based on the readings from real-world sensors. To be specific, we use the tactile readings when the robot holds the square peg without external contact, and compared between the simulation and the real world. We choose a wide range of force scaling factors and poisson convolution coefficients $c_{x,y}, \alpha_k, \beta_{ij}$, to cover the reference readings in absolute scale. We set $s_{x,y}$ and $h_{\text{max,min}}$ in the simulator to generate a similar normal force distribution. Quantitatively, we compute the MSE between the simulation forces and real tactile forces, under the no contact state when holding the square peg. Averaging over all taxels, the original simulation tactile readings have a MSE of 0.68 while the smallest MSE with \method~ over 5 random instances is only 0.11.

\noindent \textbf{Sensor Calibration.} It is inconvenient to calibrate magnet-based tactile sensors, which requires specialized hardware \cite{tomo2016uskin, tomo2018uskin2}. Moreover, manufacturers \cite{xelarobotics} will charge a significant amount of money for per-device sensor calibration. Our \method~ allows sim-to-real transfer without requiring sensor calibration process and instead learns a robust policy with large randomization.

\end{document}